\documentclass[11pt,a4paper]{article}
\usepackage[T5,T1]{fontenc}
\DeclareTextSymbolDefault{\ohorn}{T5}
\DeclareTextSymbolDefault{\uhorn}{T5}
\usepackage{arabtex}
\usepackage{utf8}
\setcode{utf8}
\usepackage[hyperref]{acl2021}

\usepackage{times}
\usepackage{latexsym}

\usepackage{booktabs}
\usepackage{xcolor}
\usepackage{xspace}
\usepackage{colortbl}
\usepackage{multirow}
\usepackage{svg}
\usepackage{amssymb}
\usepackage{enumitem}
\usepackage{arydshln}
\usepackage{comment}
\usepackage{subfig}

\definecolor{lightgreen}{RGB}{220,255,220}
\definecolor{lightblue}{RGB}{220,220,255}
\definecolor{lightred}{RGB}{255,220,220}

\newcommand{\scriptgr}{{\setlength{\fboxsep}{3pt}\colorbox{lightred}{\texttt{Script}}}}
\newcommand{\typegr}{{\setlength{\fboxsep}{3pt}\colorbox{lightblue}{\texttt{Typology}}}}
\newcommand{\bothgr}{{\setlength{\fboxsep}{3pt}\colorbox{lightgreen}{\texttt{Both}}}}
\newcommand{\script}[1]{{\setlength{\fboxsep}{3pt}\colorbox{lightred}{#1}}}
\newcommand{\typology}[1]{{\setlength{\fboxsep}{3pt}\colorbox{lightblue}{#1}}}
\newcommand{\both}[1]{{\setlength{\fboxsep}{3pt}\colorbox{lightgreen}{#1}}}

\newcommand{\F}{F$_1$\xspace}
\newcommand{\ie}{\textit{i.e.,}\xspace}
\newcommand{\eg}{\textit{e.g.,}\xspace}

\usepackage{microtype}

\aclfinalcopy 

\title{The interplay between language similarity and script \\ on a novel multi-layer Algerian dialect corpus}

\author{Samia Touileb \\
  Department of Informatics \\
  University of Oslo \\
  \texttt{samiat@uio.no} \\\And
  Jeremy Barnes \\
  Department of Informatics \\
  University of Oslo \\
  \texttt{jeremycb@uio.no} \\}

\date{}

\begin{document}
\maketitle

\begin{abstract}
Recent years have seen a rise in interest for cross-lingual transfer between languages with similar typology, and between languages of various scripts. However, the interplay between language similarity and difference in script on cross-lingual transfer is a less studied problem. We explore this interplay on cross-lingual transfer for two supervised tasks, namely part-of-speech tagging and sentiment analysis. We introduce a newly annotated corpus of Algerian user-generated comments comprising parallel annotations of Algerian written in Latin, Arabic, and code-switched scripts, as well as annotations for sentiment and topic categories. We perform baseline experiments by fine-tuning multi-lingual language models. We further explore the effect of script vs. language similarity in cross-lingual transfer by fine-tuning multi-lingual models on languages which are a) typologically distinct, but use the same script, b) typologically similar, but use a distinct script, or c) are typologically similar and use the same script. We find there is a delicate relationship between script and typology for part-of-speech, while sentiment analysis is less sensitive.
\end{abstract}

\section{Introduction}

Cross-lingual transfer has shown promising results for several tasks, however the effect of and the interplay between typologically related languages and languages that do not share the same script has seen less focus. This is especially true for under-resourced vernacular languages and dialects. 
In this paper, we focus our work on the Algerian language, a non-standardized vernacular Arabic variety, characterized by the heavy use of both code-switching and borrowings. The existing code-switching can be anything from local Algerian dialects (e.g. region based Algerian or Berber), French, English, Spanish, Modern Standard Arabic (MSA), or other Arabic dialects. The borrowings depend on the speakers' background, but is usually heavily French-based. 

Algerian is a spoken language with no standardized writing, and with the rise of social media, it has become a language extensively used to communicate online. Algerian can be written in both Arabic and Latin scripts, and code-switching can therefore occur in a mixture of scripts, or within one same script. Arabic varieties written in Latin script are referred to as Arabizi, with north African languages referred to as North African Arabizi, NArabizi in short \cite{seddah-etal-2020-building}. For the remainder of the paper, we will refer to Algerian written in Latin script as NArabizi (NA) and Algerian written in Arabic script as Algerian Arabic (DZ). 

The broad usage of Algerian results in large amounts of data, with no resources or tools to automatically process them. To address this issue and further investigate which of scripts and typological differences influence the results the most, we use a corpus of user comments that reflect the nature of the Algerian vernacular dialect: with heavy use of non-standardised spellings and code-switching.


Our main  contributions are (i) a new layer of annotations (transliteration, sentiment analysis, topic classification) that build on the Algerian NArabizi treebank corpus \cite{seddah-etal-2020-building}, (ii) we investigate the interplay of script and typology on cross-lingual transfer for the two tasks part-of-speech (POS) tagging and sentiment analysis (SA); (iii) we give a baseline model for topic 
categorization for Algerian. All of the data, annotations, and models are made freely available
\footnote{\url{https://github.com/SamiaTouileb/Narabizi}}.

To the best of our knowledge, the corpus we present in this work is the first dataset of parallel Algerian texts written in NArabizi and DZ, annotated on the morphological and syntactic levels, and for which the interplay between typology and script can be investigated. We also believe that it can help developing approaches to tackle the heavy code-switched nature of the language. 

In  what  follows,  in  Section  \ref{sec:related_work},  we give a brief overview of related works.  In Section \ref{sec:data_annotations}, we describe our dataset and annotations, the annotation processes, and give detailed statistics of the data. We start with some benchmark experiments in Section \ref{sec:benchmark}, and present in Section \ref{sec:experiments} our experiments for POS tagging, SA, and topic classification. In Section \ref{ref:results_discussion}, we summarize and discuss our results, and conclude in Section \ref{sec:conc_future} with our main findings and future plans. 

\section{Related work}\label{sec:related_work}

The vernacular Algerian language is  
under-resourced, and few freely available corpora and tools exist. Despite work  in recent years on this language \cite{adouane-etal-2020-identifying,moudjari-etal-2020-algerian,adouane-etal-2018-improving,adouane-dobnik-2017-identification,cotterell2014algerian}, there is only one corpus manually annotated for morphological and syntactical analysis  \cite{seddah-etal-2020-building}.

As pointed out by \citet{seddah-etal-2020-building}, Algerian is a non-codified spoken Semitic language. It is a morphologically-rich language \cite{tsarfaty2010statistical}, although less so than MSA \cite{saadane2015conventional}. Similarly to other north African languages, it uses heavy code-switching and borrowings, which can either be lexicalized borrowings that receive Arabic-like morphology, or borrowings that remain invariant or take the morphology of the borrowings' original language (\eg French). 
Furthermore, Algerian exhibits high variance at the morphological and phonological levels, as well as the lexicon and conventions \cite{seddah-etal-2020-building}. As shown in Table \ref{tab:lexical_var}, the Arabic name of the country ``\emph{Algeria}'' can be written in various ways in both NArabizi and DZ scripts.

\begin{table}[t]
\centering
\begin{tabular}{lr}
\toprule
NArabizi & DZ \hspace{0.7ex} \\
\cmidrule(lr){1-1}\cmidrule(lr){2-2}
al-dzayer & \RL{الدزاير}\\
dzayer &  \RL{دزاير} \\
jazayer & \RL{جزاير} \\
al-jazayer & \RL{الجزاير}\\
al-jazaair & \RL{الجزائر}\\
\bottomrule
\end{tabular}
\caption{Lexical variations of the word ``\emph{Algeria}'' in Algerian written in NArabizi and DZ scripts.}
\label{tab:lexical_var}
\end{table}

\begin{table}[t]
\centering
\begin{tabular}{llrll}
\toprule
Gloss & NArabizi & Arabic & D & Letter  \\
\cmidrule(lr){1-1}\cmidrule(lr){2-2}\cmidrule(lr){3-3}\cmidrule(lr){4-4}\cmidrule(lr){5-5}
why & we3lach & \vspace{-0.8ex} \RL{وعلاش} & 3 & \RL{ع} (ayin) 
\\

\cmidrule(lr){1-1}\cmidrule(lr){2-2}\cmidrule(lr){3-3}\cmidrule(lr){4-4}\cmidrule(lr){5-5}
he said & 9alli & \vspace{-0.8ex} \RL{قالي} & 9 & \RL{ق} (qāf)  
\\
\bottomrule
\end{tabular}
\caption{Example of non-Latin phonemes represented as digits in NArabizi. 
}
\label{tab:phonemes}
\end{table}

As in other North African languages written in Latin script, phonemes that do not exist in the Latin alphabet are represented by digits that are visually similar. For example Table \ref{tab:phonemes} shows how the digits \emph{3} and \emph{9} are used to represent the Arabic letters ``\emph{ayin}'' and ``\emph{qāf}'' respectively. 
The nature of the 
language makes it therefore an interesting avenue to explore the interplay between language similarity and differences in script on cross-lingual transfer. 

The script of NArabizi differs from the more resourceful MSA and French languages, which can be seen as its culturally closest languages. However, \citet{muller2020multilingual} show that transfer learning approaches can be used on NArabizi, both for POS-tagging and dependency parsing. They show that multilingual BERT \cite{xu-etal-2019-bert} trained on Maltese, French, and English can successfully transfer to NArabizi, despite not being included in pretraining. This shows the potential for multilingual language models to transfer to unseen dialects across scripts.  

The effect of language similarity on NLP tasks is well known \cite{ponti-etal-2019-modeling}, with several dedicated workshop series \cite{sigmorphon-2020-sigmorphon,ws-2018-nlp-similar1}. More recently, attention has turned to larger scale analyses of morphological typology effects on language modeling \cite{gerz-etal-2018-relation,cotterell-etal-2018-languages,mielke-etal-2019-kind}. Cross-lingual transfer between languages with related typology is more successful than between languages that do not share similar scripts \cite{murikinati-etal-2020-transliteration,anastasopoulos2019pushing}, especially for the study of morphological inflection. Finally, regarding difference in script, \citet{murikinati-etal-2020-transliteration} find that using high-quality transliteration as preprocessing can improve the accuracy of such models. 

\begin{table*}
\centering
    \begin{tabular}{ll}
    \toprule
    NArabizi 
    & 
    \textit{ycombati la misere li las9at fina welat kiste } \\
    \\ 
    Arabic transliteration 
    & 
    \RL{يكومباطي لا ميزار لي لسقت فينا ولات كيست } \\
    \\
    Code-switched transliteration  
    & 
    \textit{kyste} \RL{ لي لسقت فينا ولات }\textit{ la misère} \RL {يكومباطي } \vspace{1ex} \\ \cmidrule(lr){1-1}\cmidrule(lr){2-2}
    English translation 
    & he fights the misery that sticks to us and which has become a cyst \\
    \bottomrule
    \end{tabular}
    \caption{Example of transliteration annotations into Arabic and code-switched scripts. 
    The NArabizi 
    is from \cite{seddah-etal-2020-building}. 
    The translation to English is added for readers' comprehension.  }
    \label{tab:annotExample}
\end{table*}

However, in contrast to these previous works, we are interested in the interplay between similar typology and difference in script on \emph{cross-lingual transfer} for two \emph{supervised tasks}, namely POS tagging and sentiment analysis. More precisely, we are interested in investigating if there are differences in performance based on the various Algerian scripts. 

\section{Data and Annotations}\label{sec:data_annotations}

The underlying dataset we use is the NArabizi treebank presented in \citet{seddah-etal-2020-building}. This dataset comprises approximately 1,500 sentences: 1,300 NArabizi sentences extracted from an Algerian newspaper’s web forum \cite{cotterell2014algerian}, and 200 sentences from lyrics of songs collected manually from the web.
Each NArabizi sentence has five annotation layers: tokenization, morphology, identification of code-switching, syntax, and translation to French \cite{seddah-etal-2020-building}. The corpus is in \textit{conllu} format, and is freely available\footnote{\url{https://parsiti.github.io/NArabizi/}}.

To investigate the interplay between script and typology for cross-lingual transfer on 
POS tagging and SA,
we extend the annotations of \citet{seddah-etal-2020-building} by adding two levels of annotations:

\begin{description}[leftmargin=!]
    \item [\textbf{Token level:}] for each token of the NArabizi sentences we: \hfill  
    \begin{enumerate}[leftmargin=0.5cm]
        \item transliterate each NArabizi token to Arabic script (\ie DZ). 
        \item transliterate each NArabizi token to code-switched scripts (Arabic or Latin) based on the origin of the token (and the code-switch annotation label of the treebank). 
        \end{enumerate}
    \item [\textbf{Sentence level:}] we annotate each sentence of the NArabizi corpus for: \hfill
        \begin{enumerate}[leftmargin=0.5cm]
        \item sentiment: each sentence is annotated as POS (positive), NEG (negative), NEU (neutral), or MIX (a mix of two or more of the three previous classes).
        \item topic: each sentence is annotated as belonging to \emph{one} of the following topics: Politics, Prayer, Religion, Societal, Sport, or NONE.
        \end{enumerate} 
\end{description}

All the annotations were carried out by native speakers of Algerian, Arabic, and French. Two annotators worked on the token-level annotations, and three annotators for the sentence-level annotations. 
Before starting the annotations, we did a common annotation round to agree on the guidelines, and discuss possible issues. During this, we identified a set of errors in the NArabizi treebank, we therefore started by preprocessing the data and correct some of the recurring errors.
More details about our preprocessing of the dataset is given in Section \ref{sec:preprocess}, the transliteration annotations are described in Section \ref{sec:TranslitAnnot}, and sentiment and topic annotations are described in respectively Section \ref{sec:SAAnnot} and Section \ref{sec:TopicAnnot}. 

\subsection{Annotation Preprocessing}\label{sec:preprocess}

The NArabizi treebank dataset \cite{seddah-etal-2020-building} contains
duplicates both in document IDs and in sentences (strings), both across splits and within splits. Duplicate IDs refer to the same sentences, and therefore duplicate IDs imply duplicate sentences. However, duplicate sentences represent same strings with different IDs. There are far more sentence duplicates than ID duplicates. 


All duplicates were removed. However, as the corpus is already quite small, we attempt to avoid removing duplicates from the dev and test splits. If there are duplicates between the train and the dev splits, then we keep the sentences in the dev and remove them from the train set. The same is done with the test split.
For the inter-split duplicates, we identified 9 duplicated IDs and 46 (12 unique) duplicated sentences. 
%
Intra-split duplicates were only present in train split, with 9 duplicated IDs, and 28 (8 unique) duplicated sentences. We kept one occurrence of each as it seems that most of these duplicates come from the chorus of the song lyrics, and short common utterances as \eg ``viva Algeria''.

\subsection{Transliteration to Arabic and code-switched scripts}\label{sec:TranslitAnnot}

Two annotators expanded the annotations of the NArabizi treebank by \citet{seddah-etal-2020-building} by adding for each token of each sentence a transliteration into Arabic script, and a code-switched version that includes both Latin and Arabic scripts. The Latin script is used for tokens that originate from Latin-scripted languages. 

For example, Table \ref{tab:annotExample} shows how the NArabizi sentence is transliterated into the corresponding DZ and code-switched scripts. The first word, ``\<يكومباطي>'', is actually a borrowing from French. However, borrowings that are integrated into the Algerian language lexicon, and that are influenced by Arabic verbal inflections, were not written in Latin script in the code-switched annotations. 

The two annotators were given a subset of 300 sentences to transliterate, \ie these were doubly annotated. Due to the lack of codification, we do not compute any inter-annotator agreement. 
The subset of the 300 sentences were mainly used to set the annotation guidelines, and were extensively discussed by the annotators. 

We decided to normalize some of the Latin characters that do not have equivalent pronunciations in Arabic, these were transliterated into what the native annotators deemed to be the corresponding Arabic characters. In Table \ref{tab:normalised} we show the Latin letters and the Arabic form they were transliterated to. Even so, we decided to transliterate the last letter (phoneme \emph{gu}) 
into a non-native Arabic letter. This letter is vastly used in various Algerian dialects, it represents the dialectal pronunciation of \emph{qāf}, and is also used in names of places and persons. 

\begin{table}[t]
\centering
\begin{tabular}{cc}
\toprule
Letter & Transliteration  \\
\cmidrule(lr){1-1}\cmidrule(lr){2-2}
\emph{v} & \RL{ف} (\emph{f}) \\
\cmidrule(lr){1-1}\cmidrule(lr){2-2}
\emph{p} & \RL{ب} (\emph{b})  \\ 
\cmidrule(lr){1-1}\cmidrule(lr){2-2}
\emph{g} & \RL{ڨ} (\emph{gu}) \\
\bottomrule
\end{tabular}
\caption{Normalization of some of the Latin characters that do not have equivalent phonemes in Arabic.}
\label{tab:normalised}
\end{table}

We are aware of the various efforts to develop guidelines for conventional orthography of Algerian and other Arabic dialects \citep{saadane2015conventional, habash2018unified,adouane-etal-2019-normalising}, but we decided to keep the transliterations as identical as possible to the original NArabizi pronunciations and spellings, to reflect the distinctiveness of the language and its use in normal settings in social media.

During the transliteration annotations, several issues were identified in the original NArabizi treebank by \citet{seddah-etal-2020-building}. However, since our annotators were not trained to alter the dependency treebank, only a small selection of the identified errors were corrected. 

The first problem encountered is a lack of consistency in the tokenization. For example, the definite article ``\<ال>'' (
``\emph{el}'') can be found both as a stand-alone token, or attached to a word. The same applies to the adposition ``\emph{in/on}'' (``fi'' -- ``\<في>'') 
where it can be found both as a stand-alone token, and attached to the next word. For example, it was kept with the token in  ``\textit{f'doute}'' (``\textit{in doubt}''), while it was tokenized as ``\textit{f}+\textit{almarikhe}'' for the word ``\textit{falmarikhe}'' (``\textit{on Mars}''). All tokenization errors were not corrected, as this would lead to altering the dependency trees, and as previously mentioned, our annotators were not trained for this task.

Secondly, there were also errors in the translations from NArabizi to French. This is likely due to non-native Algerian speakers translating some parts of the NArabizi treebank.
We only corrected the translations that did not alter the tree, \ie the POS did not change. Some examples of these types of errors can be found in Table \ref{tab:errosTranslation} in Appendix \ref{sec:appendix}.

Finally, we also found some errors in the marker for code-switching (label \emph{lang} in the data). Some Algerian tokens were marked as French, and vice-versa. This also happened with other languages present in the data (as Spanish, English, and MSA). One of the typical errors was the acronyms of football clubs which were all labeled as Algerian. These were corrected to French, since the acronyms come from their names in French. For example the football club ``\emph{MCA}'' stands for ``\emph{Mouloudia Club d'Alger}'', while the Arabic name is 
``\<نادي  مولودية الجزائر >''
(``\emph{Nadi mouloudiat al-jazair}'').

\subsection{Sentiment annotations}\label{sec:SAAnnot}

\begin{table}[t]
    \centering
    \begin{tabular}{lrrrr}
    \toprule
    & & Train & Dev & Test \\
    \cmidrule(lr){3-3}\cmidrule(lr){4-4}\cmidrule(lr){5-5}
    \multirow{4}{*}{\rotatebox{90}{Sentiment}} & POS & 291 & 32 & 59\\
                                               & NEG & 274 & 44 & 34\\
                                               & NEU & 191 & 21 & 20\\
                                               & MIX & 242 & 40 & 31\\
    \cmidrule(lr){2-2}\cmidrule(lr){3-3}\cmidrule(lr){4-4}\cmidrule(lr){5-5}
    \multirow{6}{*}{\rotatebox{90}{Topic}}     & NONE     & 300 & 34 & 36\\
                                               & Politics & 80  & 11 & 16\\
                                               & Prayer   & 38  & 9  & 9\\
                                               & Religion & 17  & 4  & 1\\
                                               & Societal & 204 & 25 & 31\\
                                               & Sport    & 359 & 54 & 51\\
    \bottomrule
    \end{tabular}
    \caption{Distribution of sentiment and topic annotations.}
    \label{tab:distSA}
\end{table}

The sentences were classified based on their polarities into four different classes: POS (positive), NEG (negative), NEU (neutral), and MIX (mixed). The annotation guidelines were quite simple, and annotators were asked to use POS and NEG in clear positive and negative cases respectively. If a sentence does not express any kind of polarity, then NEU was assigned. When sentences express a combination of two or more of the POS, NEG, or NEU polarities, annotators were asked to assign the MIX label. The inter-annotator agreement using \emph{Cohen's kappa coefficient} \(\kappa \) is 0.71 on the doubly annotated subset of 300 sentences.
Table \ref{tab:distSA} shows the distribution of the four labels across the training, development, and test sets. The distribution is unbalanced, and the large amount of sentences categorized as MIX can be problematic as it can contain all other polarities. However, the difference between the POS and NEG classes is relatively small, which we believe should be suitable for binary sentiment classification tasks.

\subsection{Topic annotations}\label{sec:TopicAnnot}


After a first round of common analysis in collaboration with the annotators, we identified five topics. However, 
some sentences were difficult to classify 
and we therefore decided to include the category ``\emph{NONE}''. The final dataset is annotated for the following six categories: (1) \emph{Politics}: contains all sentences referring or discussing political events or issues; (2) \emph{Prayer}: all sentences representing prayers; (3) \emph{Religion}: sentences discussing religious issues or issues related to religion in general; (4) \emph{Societal}: societal related discussions. Covers everything from schools and teaching, to terrorism and extremism; (5) \emph{Sport}: mainly covering football events, but spans all types of sports and related events; (6) \emph{NONE}: sentences that were impossible to categorize. This was mainly due to the lack of context, as some sentences were comments responding to either articles or other comments. The final \(\kappa \) score for the triply annotated 300 sentences was 0.70.

Table \ref{tab:distSA} also shows the distribution of topics across the three splits. Most sentences were classified as ``Societal'' and ``Sport''. A large amount of sentences could not be categorised, and few sentences were related to ``Religion'' and ``Prayer''. Due to the size of the two latter, one could argue that they could be collapsed into a single topic, as done in our benchmarking experiments (see Section \ref{sec:benchmark}). However, we decided to keep them separate in the annotations, to facilitate further annotations in the future. 

\section{Benchmarking experiments}\label{sec:benchmark}

\begin{table}[]
    \centering
    \begin{tabular}{llrrr}
    \toprule
            && NA & DZ & CS \\
            \cmidrule(lr){3-3}\cmidrule(lr){4-4}\cmidrule(lr){5-5}
    \multirow{4}{*}{\rotatebox{90}{Sentiment}}  & BOW   & 47.5 & 45.1 & \textbf{49.5} \\
               & AVE & 52.5 & 43.2 & 36.7 \\
               & CNN & 50.2 & \textbf{50.4} & 46.2\\
               & BiLSTM & \textbf{53.9} & 45.9 & 45.6\\
               \cmidrule(lr){1-2}\cmidrule(lr){3-3}\cmidrule(lr){4-4}\cmidrule(lr){5-5}
    \multirow{4}{*}{\rotatebox{90}{Topic}} & BOW & 25.8 & 34.9 & \textbf{38.1}\\
          & AVE & 40.9 & 44.6 & 22.8\\
          & CNN & 24.4 & 33.4 & 27.4 \\
          & BiLSTM & \textbf{49.4} & \textbf{57.0} & 36.4 \\
    \bottomrule
    \end{tabular}
    \caption{Benchmark results for Sentiment Analysis and Topic classification on the three varieties of the dataset: NArabizi (NA), Algerian Arabic (DZ), and a code-switched version (CS). Sentiment and Topic are both Macro \F.}
    \label{tab:benchmark}
\end{table}

\begin{table*}[]
    \centering
    \begin{tabular}{llrrrrrr}
    \toprule
    & & \multicolumn{2}{c}{UPOS} & \multicolumn{4}{c}{Sentiment} \\
    \cmidrule(lr){3-4}\cmidrule(lr){5-8}
   group & language & \# sents. & avg. len. &  \# docs. & avg. len. & pos & neg \\
    \cmidrule(lr){1-1}\cmidrule(lr){2-2}\cmidrule(lr){3-3}\cmidrule(lr){4-4}\cmidrule(lr){5-5}\cmidrule(lr){6-6}\cmidrule(lr){7-7}\cmidrule(lr){8-8}
    
Original & NArabizi & 1,276 & 16.1 & 731 & 14.4 & 380 & 351 \\
\scriptgr & Persian & 5,997 & 26.3 & 879 & 49.6 & 419 & 460\\
\scriptgr & Urdu    & 5,130 & 27.0 & 980 & 17.5 & 480 & 500  \\
\typegr & Hebrew    & 6,216 & 30.6 & 12,434 & 24.0 & 8,512 & 3,922 \\
\typegr & Maltese   & 2,074 & 21.8 & 719 & 18.7 & 237 & 482 \\
\bothgr & MSA       & 7,664 & 42.3 & 51,051 & 60.8 & 42,828 & 8,223 \\
    \bottomrule
    \end{tabular}
    \caption{Statistics of UPOS and binarized sentiment data.}
    \label{tab:PosStats}
\end{table*}

We perform benchmark experiments for SA and topic classification. Specifically, we use the setup from \newcite{barnes-etal-2017-assessing}, who perform experiments with a logistic regression classifier with  bag-of-words features (BOW) and averaged embedding features (AVE), as well as a CNN and BiLSTM. We use their default value for hyperparameters (c=1, hidden dimension = 100, dropout = 0.3) and train for 20 epochs, finally testing the best model on the dev set. As the label distribution for both tasks is highly skewed, we use Macro \F to evaluate. Given the size of the categories ``Prayer'' and ``Religion'', we collapse them to a single topic, converting the topic classification task into a 5-class multi-class problem. 

For the NArabizi and code-switched experiments, we create 100-dimensional fasttext embeddings \cite{bojanowski-etal-2017-enriching} on the unlabeled NArabizi data made available by \newcite{seddah-etal-2020-building}. For the DZ experiments, we use available 300-dimensional MSA fasttext embeddings\footnote{Available at \url{https://fasttext.cc/docs/en/pretrained-vectors.html}.} trained on Wikipedia articles.

Table~\ref{tab:benchmark} shows the results. On the sentiment task, the BiLSTM performs best on NArabizi, the CNN best on DZ, and BOW best on the code-switched data. For topic classification the performance is similar, but BiLSTM is also best on DZ. The fact that BOW performs best on code-switched data is largely due to the large amount of out-of-vocabulary words for all other methods, which require embeddings. These baseline experiments show that the dataset is challenging, and the variation means that no single model is always best. The code-switched setting is particularly challenging.

\section{The interplay between language similarity and script}\label{sec:experiments}

The transliteration and further sentiment and topic annotations allow us to explore what interplay there is between typology and script in cross-lingual transfer. \newcite{muller2020multilingual} perform experiments on zero-shot cross-lingual transfer for POS tagging on NArabizi. They find that the best transfer language is Maltese, a Semitic language which is written in Latin script, rather than MSA, which performs poorly. This begs the question: \emph{is it mainly similar typology or a shared script that leads to this result?} %
The transliterated dataset, along with the further sentiment annotations, allow us to investigate this question in more depth, as we are able to control for the script choice.

We choose Persian and Urdu, languages written in Arabic script, but morphologically distinct from DZ (we refer to this group as \scriptgr), Hebrew and Maltese, two Semitic languages written in other scripts (\typegr) , and MSA, which is both morphologically similar and written in Arabic script (\bothgr). These languages are both available in UD \cite{UD2.6} and also have available sentiment analysis datasets (Hebrew \cite{amram-etal-2018-representations}, Maltese \cite{Dingli2016}, MSA \cite{nabil-etal-2015-astd, abdulla-etal-2013}, Urdu \cite{khan-2020}, Persian \cite{hosseini-etal-2018}). As not all sentiment datasets have the same labels as the NArabizi dataset, we remove all neutral and mixed labels and create binary sentiment data for all languages.

Table \ref{tab:PosStats} gives an overview of the statistics of the POS and SA datasets, respectively. The NArabizi data is the smallest POS data (1,276 sentences), followed by Maltese (2,074), Urdu (5,130), Persian (5,997), Hebrew (6,216), and finally MSA (7,664). The average sentence lengths in tokens range between 16.1 for NArabizi and 42.3 in MSA. The sentiment datasets have a larger variance, ranging from 719 sentences for Maltese to 51,051 for MSA. The distribution of polarity is also skewed to a different degree in each dataset.

\subsection{Modeling}

We model universal POS (UPOS) tagging as a sequence labeling task and SA as a classification task using 
multilingual BERT \cite{xu-etal-2019-bert}.
We fine-tune each model on the available training data in each language, using a shared set of hyperparameters which were selected from recommended values according to the characteristics of our data. We set the learning rate to 2e-5, max sequence length of 256, batch size of 8 or 16\footnote{Depending on the size of the training set, model architecture, and available GPU memory.}, and perform early stopping once the validation score has not improved in the last epochs, saving the model that performs best on the dev set. We then test each model on its own dev and test data, the NArabizi test set, and finally the transliterated data. We use accuracy as our metric for POS and macro \F for sentiment, as the latter often contains unbalanced classes, and define a baseline as the result of predicting the majority class.

\section{Results and Discussion}\label{ref:results_discussion}

In order to quantify the zero-shot loss, we define a measure of average transfer loss between a group in  Equation \ref{eq:transfer_formula}:
\begin{equation}\label{eq:transfer_formula}
    TL_{x \rightarrow y} = S_{x \rightarrow x} - S_{x \rightarrow y}
\end{equation}
where $TL_{x \rightarrow y}$ is the transfer loss experienced by a model fine-tuned in language $x$ when transferring to language $y$ and $S_{x \rightarrow y}$ is the score\footnote{The score metric will depend on the task: accuracy in POS and macro \F in sentiment analysis.} achieved when testing a model fine-tuned in language $x$ on language $y$. Thus, it is a measure of the performance lost in the transfer process.

We also define its averaged variant:
\begin{equation}
    \overline{TL}_{A \rightarrow B} = \frac{1}{N_A} \sum_{i \in A} \overline{TL}_{i \rightarrow B}
\end{equation}
where $\overline{TL}_{A \rightarrow B}$ refers to the average transfer loss experienced by languages from any group $A$ to languages from group $B$ (\textit{group-to-group} transfer loss) and $N_A$ is the number of languages included in the experiment that belong to group $A$ (in our case, either languages that have similar typology, or have the same script).

\subsection{POS}
Table~\ref{tab:cross_pos} shows the results for the POS tagging. For completeness, we compare with the results from \newcite{seddah-etal-2020-building}, who use a feature-based alVWTagger, described in more detail in \newcite{de-la-clergerie-etal-2017-parisnlp} and \newcite{muller2020multilingual}, who use mBERT and the StanfordNLP tagger \cite{qi-etal-2018-universal}. 

\begin{table}[]
    \centering
\begin{tabular}{llrrrr}
\toprule
              & Dev & Test & NA & DZ \\
\cmidrule(lr){2-2}\cmidrule(lr){3-3}\cmidrule(lr){4-4}\cmidrule(lr){5-5}\cmidrule(lr){6-6}
Maj. & -- & -- & 19.9 & 19.9 \\
1-NArabizi & & -- & 80.4 & --\\
                 2-NArabizi &            -- & -- & 81.6 & -- \\
                  2-Maltese   &     -- & -- & 35.1 & -- \\
\cmidrule(lr){1-1}\cmidrule(lr){2-2}\cmidrule(lr){3-3}\cmidrule(lr){4-4}\cmidrule(lr){5-5}\cmidrule(lr){6-6}
  NArabizi & 77.1 & -- & 76.3 &  43.6   \\
  Algerian (DZ) &  83.2 & -- & 39.9  & 82.5  \\
  \cmidrule(lr){1-1}\cmidrule(lr){2-2}\cmidrule(lr){3-3}\cmidrule(lr){4-4}\cmidrule(lr){5-5}\cmidrule(lr){6-6}
                  \multirow{1}{*}{\script{Persian}}  & 95.8 & 95.5 & 22.7 & 26.5 \\
                    \multirow{1}{*}{\script{Urdu}}  &  94.0 & 93.4 & 18.7 & 21.6 \\
                \multirow{1}{*}{\typology{Hebrew}} & 97.4    & 96.8 & 32.7 & 38.2 \\
                 \multirow{1}{*}{\typology{Maltese}} &  93.6 & 93.0 &    \textbf{37.8} &     \textbf{38.4} \\
                  \multirow{1}{*}{\both{MSA}} &     97.0 & 96.7 &     20.0 &     30.5 \\
\bottomrule
\end{tabular}
    \caption{POS accuracy when training on Train Lang. Dev Acc. and Test are in-language, while Test Acc. on Narabizi and Algerian (DZ) is zero-shot cross-lingual. 1-\newcite{seddah-etal-2020-building}, 2-\newcite{muller2020multilingual}.}
    \label{tab:cross_pos}
\end{table}

\begin{figure}
    \centering
    \includegraphics[scale=0.45]{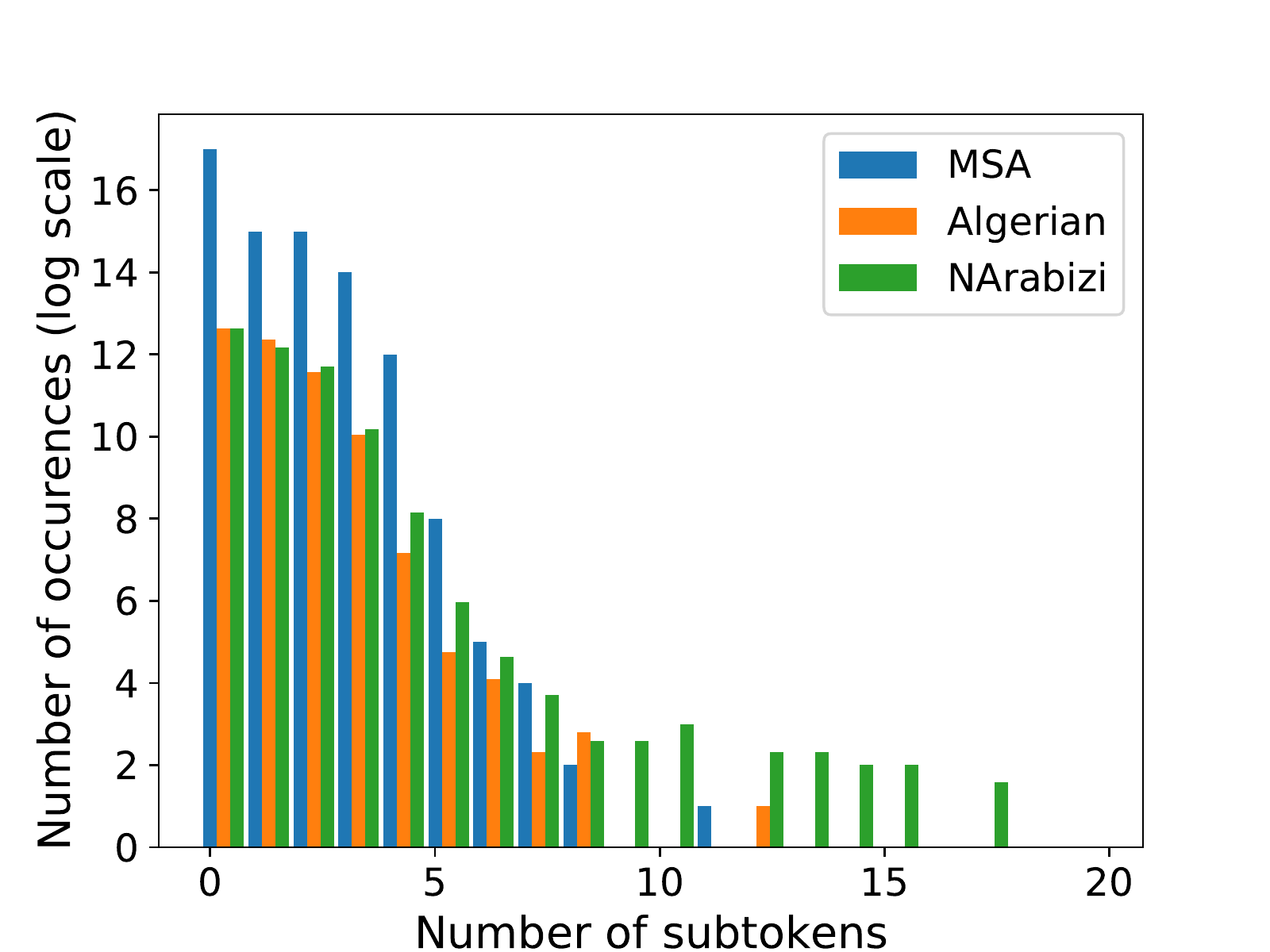}
    \caption{The effect of mBERT tokenization on MSA, Algerian Arabic (DZ), and NArabizi.}
    \label{fig:tokenization}
\end{figure}

Hebrew has the best test 
accuracy (96.8)
and Maltese the worst 
(93.8), while the others are somewhere between. 
All models perform better on the transliterated data than the original NArabizi, although 
training on Urdu performs lower than the majority baseline. This suggests that even though mBERT was not pretrained on NArabizi or DZ, there is a preference for DZ. This is likely due to the fact that at least \emph{some} of the words have been seen in pretraining, \ie through MSA. An analysis of the tokenization shows that mBERT splits NArabizi words at a much higher rate than DZ (see Figure \ref{fig:tokenization}), breaking it into smaller pieces, which may account for some of the differences between the two. The fact that training on Maltese achieves the best score on both NArabizi (37.8) and DZ (38.4), however, suggests that there is still an effect of typology. 

The monolingual model trained and tested on DZ performs better (82.5 acc.) than the one trained and tested on NArabizi (76.3). When each of these models is tested on the other, they have significant transfer losses (32.7 for NArabizi~$\rightarrow$~DZ, and 42.6 for DZ~$\rightarrow$~NArabizi). Here too, transfer to DZ script seems easier.

On POS, the effect of language typology is stronger than script, with the best results achieved by training on Maltese and Hebrew. The average transfer loss from Persian and Urdu to DZ is 
70.4 while for Hebrew and Maltese to NArabizi it is 
59.6, showing less transfer loss from \typegr. MSA has higher transfer loss on NArabizi 
(76.7) than DZ 
(66.2).
The differences between average transfer loss on NArabizi and DZ are also slightly larger for \scriptgr 
(3.4) compared to \typegr (3.1) or MSA (3.1). 

All of this points to a complicated relationship between script and typology on POS. First of all, it is clear that mBERT prefers the Arabic script seen in pretraining. At the same time, typological similarity also plays a strong role in cross-lingual transfer in POS, although even in this case, the best scores are found on DZ.

\subsection{Sentiment}

Table~\ref{tab:cross_sent} shows the results for sentiment analysis. Training in-language again produces the best results (72.1 and 80.3 on NArabizi and DZ, respectively). The transfer loss from NArabizi to DZ is relatively low (9.0), while inversely it is immense (52.6).

Like on POS, most models perform better on DZ and the best zero-shot results do not come from training on \typegr. In fact, quite the opposite, as these lead to the worst scores and have the highest average transfer loss (34.8/27.9). The best models are MSA for NArabizi (62.4) and Urdu for DZ (63.9), which curiously performs better than NArabizi $\rightarrow$ DZ. MSA has transfer losses of 12.8/25.1, while \scriptgr have the lowest average transfer loss (9.2/4.2). This suggests that cross-lingual transfer for a more semantic task, \eg sentiment analysis, is less reliant on both typological and script similarities.

\begin{table}[]
    \centering
\begin{tabular}{lrrrr}
\toprule
              & Dev & Test & NA & DZ \\
\cmidrule(lr){1-1}\cmidrule(lr){2-2}\cmidrule(lr){3-3}\cmidrule(lr){4-4}\cmidrule(lr){5-5}
 Maj. & -- & -- & 39.0 & 39.0 \\
  NArabizi &    78.8 & --  &   72.1 &     63.1 \\
  Algerian (DZ) &    84.9 & --  &  27.7 &     80.3 \\
  \cmidrule(lr){1-1}\cmidrule(lr){2-2}\cmidrule(lr){3-3}\cmidrule(lr){4-4}\cmidrule(lr){5-5}
   \script{Persian} &    65.9 & 66.2 &   56.9 &     56.2 \\
      \script{Urdu} &    59.0 & 62.4  &  53.3 &     \textbf{63.9} \\
    \typology{Hebrew} &    88.4 & 88.7 &   47.2 &     52.2 \\
   \typology{Maltese} &    63.7 & 61.8  &  33.8 &     42.5 \\
    \both{MSA} &    74.2 & 75.2  &  \textbf{62.4} &     50.1 \\

\bottomrule
\end{tabular}
    \caption{Macro \F on the zero-shot cross-lingual sentiment task. Note that these results are not comparable to the benchmark experiments, as the data has been converted to binary sentiment classification in order to perform the cross-lingual experiments.}
    \label{tab:cross_sent}
\end{table}

\subsection{Analysis of results}

\begin{figure*}%
    \centering
    \subfloat{{\includegraphics[width=.45\textwidth]{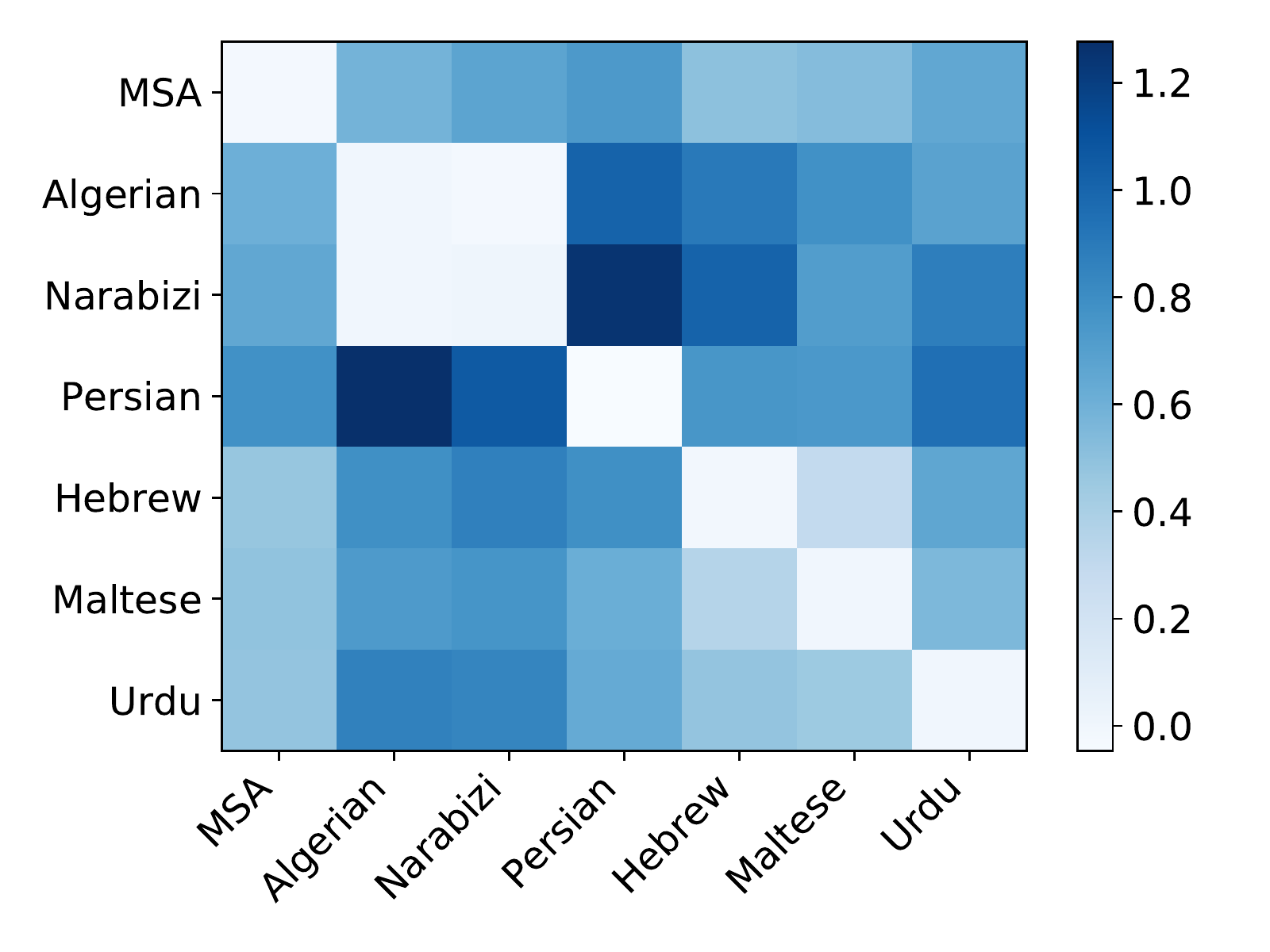} }}%
    \qquad
    \subfloat{{\includegraphics[width=.45\textwidth]{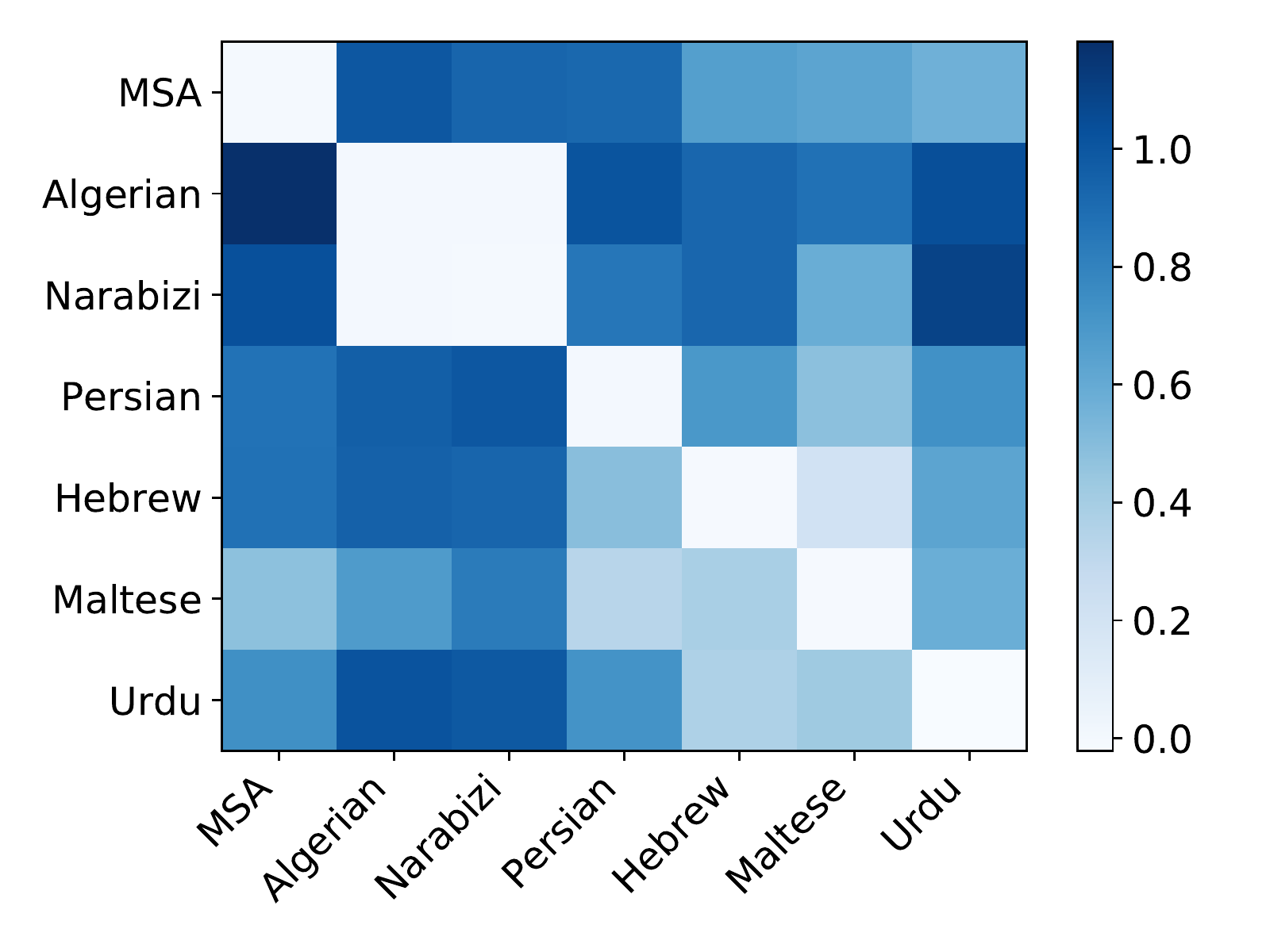} }}%
    \caption{Pairwise proxy A distance between English translations of sentiment (left) and POS (right) datasets.}%
    \label{fig:proxyA}%
\end{figure*}

As domain differences between datasets could also lead to transfer loss, we control for this variable by first translating all data to English (to use as a pivot language) and calculating domain difference using Proxy A-distance. Proxy A-distance \cite{Glorot2011} measures the generalization error of a linear SVM trained to discriminate between two domains. We translate 1,000 sentences from each dataset to English using GoogleTranslate and then compute the proxy A-distance\footnote{Implementation adapted from the code available at \url{https://github.com/rpryzant/proxy-a-distance}.} We show heat maps for the domain distances in Figure \ref{fig:proxyA}.

For POS tagging, there are small but insignificant negative effects of proxy A-distance on results (a Pearson coefficient of -0.264, $p>0.05$). On the sentiment task, there is no significant domain effect (0.264, $p>0.05$). This suggests that most of the transfer loss is not due to domain mismatch.

\section{Conclusion and Future work}\label{sec:conc_future}

In this paper we have described the process of annotating an available Algerian corpus with sentiment and topics, as well as the transliteration to Arabic and code-switched scripts, and finally some aspects of corpus cleanup. We performed benchmark experiments on the three script varieties and show that they are a challenging testbed for future experiments.

We used this new resource to explore a valuable research question in cross-lingual transfer: namely, what is the interplay between language similarity and script when choosing a source language? 
We found there is a delicate interplay between similar typology and script for transfer in part-of-speech tagging, where typology is more important, but having seen the script in pretraining also influences results. Sentiment analysis, on the other hand, is less sensitive to typological differences, while still preferring the script seen in pretraining. This suggests that choice of transfer language is task-specific and that surprising differences can appear from one task to another.

In the future, we would like to address data related issues, and correct the tokenization and translation issues discussed in Section \ref{sec:preprocess}. Moreover, we plan to focus more concretely on the code-switching aspect of our dataset. The challenges of code-switched data to NLP techniques are numerous, and we would like to focus on the syntactic analysis of our code-switched data, and to explore in more details language modeling approaches to processing it. 

\bibliographystyle{acl_natbib}
\bibliography{anthology,acl2021}

\appendix
\section*{}

\newpage
\section*{}

\newpage
\section{Appendix}\label{sec:appendix}

\vspace{2em}
\begin{minipage}{\textwidth}
    \centering
    \begin{tabular}{
    p{0.25\textwidth}
    p{0.1\textwidth}
    p{0.13\textwidth}
    p{0.25\textwidth}
    c}
    \toprule
    id -- Sentence (En) & Translation & Correct (En) & Explanation &  Status \\
    \cmidrule(lr){1-1}\cmidrule(lr){2-2}\cmidrule(lr){3-3}\cmidrule(lr){4-4}\cmidrule(lr){5-5}
    1 -- \textit{Mabrouk ya \underline{lafhal}} 1 (congratulations oh brave) & Lafhal & courageux (brave) & annotated as PROPN, should be NOUN. & X  \\
    \cmidrule(lr){1-1}\cmidrule(lr){2-2}\cmidrule(lr){3-3}\cmidrule(lr){4-4}\cmidrule(lr){5-5}
    2 -- \textit{el hamdou lilah ya rabi \underline{alla} 3awdat chawchi} (thanks God for the return of Chawchi) & Allah & alla (for) & \vspace{-2.2ex} this is the word \< على > annotated as PROPN, should be DET. & X \\
    \cmidrule(lr){1-1}\cmidrule(lr){2-2}\cmidrule(lr){3-3}\cmidrule(lr){4-4}\cmidrule(lr){5-5}
    3 -- \textit{vive toi \underline{mbolhi}} (long live you Mbolhi) & fou (crazy) & Mbolhi & Mbolhi is the name of a football player. It is not an ADJ, should be PROPN. & X\\
    \cmidrule(lr){1-1}\cmidrule(lr){2-2}\cmidrule(lr){3-3}\cmidrule(lr){4-4}\cmidrule(lr){5-5}
    4 -- \textit{\underline{mabka} fiha ghure sehab elderaham} (the only ones remaining are those with money) & pas pleurer (not cry) & mabka (only remain) & \vspace{-2.2ex}  this is the word \<ما بقى> (only remain) and not \<ما بكى> (not cry). & \checkmark \\
     \cmidrule(lr){1-1}\cmidrule(lr){2-2}\cmidrule(lr){3-3}\cmidrule(lr){4-4}\cmidrule(lr){5-5}
    5 -- \textit{al \underline{mou3ak} fil jazair mayakdarch yakhrouj} (the handicapped in Algeria can't go out) & obstacle (obstacle) & handicapé (handicapped) & the word \textit{mou3ak} in this context means handicapped. & \checkmark\\
    
    \bottomrule
    \end{tabular}
    \captionof{table}{Examples of errors present in the NArabizi treebank. Status ``X'' means not corrected, while status ``\checkmark'' means corrected.}
    \label{tab:errosTranslation}
\end{minipage}

\end{document}